\documentclass{article}
\usepackage[table, dvipsnames, svgnames, x11names]{xcolor}

\usepackage[T1]{fontenc}
\PassOptionsToPackage{sort&compress}{natbib}
\usepackage[preprint]{corl_2026}

\usepackage{booktabs}
\usepackage{graphicx}
\usepackage{amsmath}
\usepackage{array}
\usepackage{capt-of}
\usepackage{tikz}
\usetikzlibrary{arrows.meta,decorations.pathreplacing,positioning,shadows}

\definecolor{cR}{HTML}{D55E00}
\definecolor{cB}{HTML}{0072B2}
\definecolor{cG}{HTML}{009E73}
\definecolor{cY}{HTML}{E69F00}
\definecolor{cP}{HTML}{CC79A7}
\definecolor{cK}{HTML}{8C8C8C}
\tikzset{cone/.style={fill=black!8}}

\newcommand{\posthoc}{\textsuperscript{$\dagger$}}

\title{Grid or Nodes? Probing Object Memory\\in World Models Under Partial Observability}

\author{
  Justin McMillen\\
  University of South Florida\\
  Tampa, FL, USA\\
  \texttt{jmcmillen@usf.edu}
  \And
  Yasin Yilmaz\\
  University of South Florida\\
  Tampa, FL, USA\\
  \texttt{yasiny@usf.edu}
}

\begin{document}
\maketitle
\hypersetup{pdfsubject={}}

\begin{abstract}
A robot's world model must remember objects that leave the camera's view. FloWM, a learned world model, keeps them in a spatial grid of features that it moves with five built-in velocities. A node memory instead keeps each object's position and velocity as numbers. We compare them in FloWM's own block world, with an oracle-fed Kalman tracker as the node memory. How the grid is read changes its score: a single-cell probe finds 66\% of out-of-view blocks, while FloWM's own probe 89\%. FloWM also scores 8 points higher in view although no frame is shown, which we attribute to its write-back of its own predictions. Out of view, the tracker reaches FloWM's range in 14KB against 5.6MB, with no probe. On diagonal motion, outside its five velocities, FloWM finds 11\% of blocks even after fine-tuning, while the unchanged tracker keeps 93\%. Faster blocks defeat both until one more allowed speed takes the tracker to 94\%, where FloWM's built-in shifts would need 25 copies of its grid. A learned node memory must still replace the oracles and handle velocity under noise, long gaps, and the bounce.

\end{abstract}


\section{Introduction}
\label{sec:intro}

A robot that turns away from a moving cart must still know where the cart has gone. Robots increasingly plan with learned world models~\citep{hafner2025dreamerv3,zhou2025dinowm,assran2025vjepa2}, so these models need a memory for objects that leave the camera's view. Where should that memory keep a moving object? FloWM~\citep{lillemark2026flowm} keeps a top-down grid of learned features and moves it with a fixed set of built-in velocities. Its motion prior, what it assumes about motion, is therefore part of the memory's layout. Our node-based memory platform instead stores one node per object, with its position and velocity as numbers.

Other video world models add a memory for consistency when the camera returns~\citep{xiao2025worldmem,huang2025memoryforcing}, or keep state for subjects out of view~\citep{duan2026liveworld,chen2026hydra}. Related to a node memory are tracking by detection~\citep{bewley2016sort}, models of object permanence~\citep{shamsian2020permanence,tokmakov2021permatrack} and object-centric models with one representation per object~\mbox{\citep{locatello2020slot,kipf2020cswm,creswell2021oat,daniel2026lpwm,nishimoto2026stica}}. Memories are usually judged by probes, networks trained to read out a property~\citep{li2023othello,gregor2019shaping,wijmans2023emergence}, whose scores reflect the probe as well as the memory~\citep{hewitt2019control,belinkov2022probing}. Benchmarks test what world models \mbox{remember}~\citep{lu2026wrbench,ma2026stevo,chen2026memobench,gu2026r2mbench,lian2025loopnav}, and others question what replay quality or next-step accuracy shows~\citep{king2026echomemory,vafa2024evaluating,zhang2026worldinworld}.

We compare the two designs on FloWM's own task, a room of sliding blocks. Each memory receives 70 steps of context and then predicts 70 steps from the agent's actions alone. FloWM works from pixels, and the node memory is a Kalman tracker with oracle inputs: the simulator gives it block masks and slightly noisy positions. The tracker as-is is not a deployable system.

Does the grid hold a block's position, or does the probe compute it (Sec.~\ref{sec:q1})? On FloWM's own setting, how close does a node memory come (Sec.~\ref{sec:q2})? What happens outside the built-in velocities (Sec.~\ref{sec:q3})? What stands between the tracker and a full platform? (Sec.~\ref{sec:q4})?

\section{Two memories}
\label{sec:memories}

\subsection{Environment and protocol}
\label{sec:env}

\begin{figure}[t]
\centering

\begin{tikzpicture}[
    font=\scriptsize,
    blk/.style={minimum size=5.5pt, inner sep=0pt, draw=black!70, line width=0.3pt},
    mv/.style={-{Stealth[length=3pt]}, line width=0.6pt},
    cell/.style={draw=black!35, line width=0.2pt},
    box/.style={draw=black!60, rounded corners=2pt, align=center, inner sep=1.4pt, fill=white,
                line width=0.4pt, text width=2.65cm},
    nbox/.style={draw=black!60, rounded corners=2pt, align=center, inner sep=1.4pt, fill=white,
                 line width=0.4pt, text width=1.9cm},
    card/.style={draw=black!60, rounded corners=2pt, fill=white, line width=0.4pt,
                 minimum width=2.2cm, minimum height=0.62cm, anchor=west, inner sep=0pt},
    flow/.style={-{Stealth[length=3.5pt]}, line width=0.5pt, black!70},
    lab/.style={anchor=north west, font=\small\bfseries, inner sep=0pt},
  ]


  \begin{scope}
    \fill[cone] (0.9,1.1) -- ++(30:1.4) arc (30:-30:1.4) -- cycle;
    \draw[line width=0.8pt] (0,0) rectangle (2.45,2.45);
    \fill[black] (0.9,1.1) ++(0:0.13) -- ++(150:0.18) -- ++(-90:0.18) -- cycle;
    \node[blk, fill=cR] (r) at (0.45,2.0) {};
    \draw[mv] (r.east) -- ++(0.4,0);
    \node[blk, fill=cB] (b) at (0.4,0.75) {};
    \draw[mv] (b.south) -- ++(0,-0.35);
    \node[blk, fill=cG] (g) at (2.05,1.2) {};
    \draw[mv] (g.north) -- ++(0,0.35);
    \node[blk, fill=cY] (y) at (1.85,0.28) {};
    \draw[mv] (y.west) -- ++(-0.4,0);
    \node[blk, fill=cP] (p) at (0.28,1.4) {};
    \draw[mv, dashed] (p.north east) -- ++(0.27,0.27);
    \node[blk, fill=cK] (k) at (2.2,2.22) {};
    \draw[-{Stealth[length=3pt]Stealth[length=3pt]}, line width=0.6pt, dashed] (k.west) -- ++(-0.7,0);
    \node[anchor=north west, inner sep=1pt] at (0.8,1.02) {agent};
    \node[lab] at (0,2.95) {(a)};
  \end{scope}


  \begin{scope}[xshift=2.85cm]

    \foreach \k in {4,3,2,1,0} {
      \begin{scope}[shift={(0.1*\k, 0.1*\k)}]
        \fill[white] (0,0.6) rectangle (1.2,1.8);
        \draw[cell, step=0.2] (0,0.6) grid (1.2,1.8);
        \draw[black!70, line width=0.4pt] (0,0.6) rectangle (1.2,1.8);
      \end{scope}
    }
    \fill[cR] (0.4,1.0) rectangle (0.6,1.2);
    \draw[mv, cR] (0.62,1.1) -- ++(0.35,0);
    \node[align=center, anchor=north] at (0.6,0.55) {5 grid copies:\\still, $\pm x$, $\pm y$};

    \node[box, anchor=north west] (s1) at (1.72,2.55) {move copies by their velocity};
    \node[box, below=0.8mm of s1] (s2) {move against agent motion};
    \node[box, below=0.8mm of s2] (s3) {mix copies (3$\times$3 conv)};
    \node[box, below=0.8mm of s3] (s4) {render in-view cells};
    \node[box, below=0.8mm of s4, draw=black, line width=0.9pt] (s5) {\textbf{write in-view cells}};
    \node[anchor=north west, inner sep=1pt, text=black!70] at ([yshift=-1.2mm]s5.south west) {in prediction it writes its own render};
    \draw[flow] (s1) -- (s2);
    \draw[flow] (s2) -- (s3);
    \draw[flow] (s3) -- (s4);
    \draw[flow] (s4) -- (s5);
    \draw[flow] (s5.east) -- ++(0.15,0) |- (s1.east);
    \draw[decorate, decoration={brace, amplitude=2.5pt}, black!50] ([xshift=3mm]s1.north east) -- ([xshift=3mm]s2.south east)
      node[midway, right=2.5pt, align=left, text=black!70] {built\\in};
    \draw[decorate, decoration={brace, amplitude=2.5pt}, black!50] ([xshift=3mm]s3.north east) -- ([xshift=3mm]s5.south east)
      node[midway, right=2.5pt, align=left, text=black!70] {learned};
    \node[lab] at (-0.1,2.95) {(b)};
  \end{scope}


  \begin{scope}[xshift=8.95cm]
    \foreach \col/\y in {cR/2.24, cB/1.52, cG/0.80} {
      \node[card] at (0,\y) {};
      \node[blk, fill=\col] at (0.2,\y) {};
      \node[anchor=west, inner sep=0pt, align=left] at (0.42,\y) {$x,\,y,\;v_x,\,v_y$\\uncertainty, color};
    }
    \node[align=center, anchor=north] at (1.1,0.3) {one node per block};
    \node[align=center, anchor=north, text=black!70] at (3.55,0.3) {hand-set, oracle inputs};

    \node[nbox, anchor=north west] (n1) at (2.55,2.55) {move each node,\\snap speed, bounce};
    \node[nbox, below=1.6mm of n1] (n2) {match detections\\to nodes};
    \node[nbox, below=1.6mm of n2] (n3) {update matched\\nodes};
    \draw[flow] (n1) -- (n2);
    \draw[flow] (n2) -- (n3);
    \draw[flow] (n3.east) -- ++(0.15,0) |- (n1.east);
    \node[lab] at (-0.1,2.95) {(c)};
  \end{scope}

\end{tikzpicture}
\caption{(a)~The room from above. Solid arrows show motion in the standard setting, and dashed arrows the diagonal and speed-2 (double head) motion that the varied setting adds. (b)~One prediction step of FloWM's grid memory. (c)~The tracker, a node memory.}
\label{fig:setup}
\end{figure}
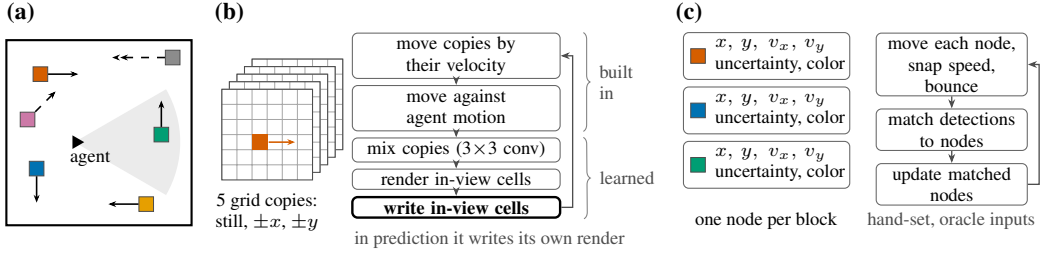

Dynamic Block World~\citep{lillemark2026flowm} is a Miniworld~\citep{chevalierboisvert2023minigrid} room (Fig.~\ref{fig:setup}a) whose floor is a grid of cells. Each evaluation video holds six blocks, one per color, which slide across the floor and bounce off the walls.
An agent moves one cell or turns $90^\circ$ per step and sees through a $60^\circ$ camera. Each video has a \emph{context phase} of 70 frames with actions and a \emph{prediction phase} of 70 actions without frames.

At every predicted step, each memory names, for each block, one cell of a grid centered on the agent. Only the block's exact cell is correct, so chance is below 0.1\%. A \emph{sighting} is a context frame in which the block is visible. No frame is shown during prediction. We call a block \emph{in view} at a step when it lies inside the field of view the camera would have then, and \emph{out of view} otherwise.

In the \emph{standard setting}, which follows FloWM's published data, every block moves one cell per step along one floor axis. In our \emph{varied setting}, each block draws a random nonzero velocity with whole-number components from $-2$ to $2$ cells per step. Only 17\% of blocks then move as in the standard setting (\emph{built-in}), and 16\% move one cell per step on both axes (\emph{diagonal}). The other 67\% move two cells per step on some axis (\emph{speed 2}). We tested on 256 standard and 1,024 varied videos, and fitted probes and tracker settings on separate tuning and training videos (Apps.~\ref{app:protocol} and~\ref{app:tracker}).

\subsection{FloWM's grid memory}
\label{sec:flowm}

FloWM is trained end to end to predict camera frames.
Its memory is a top-down grid around the agent, $33\times33$ cells that each hold 256 learned features. The memory holds five copies of this grid, one per built-in velocity: standing still, and one cell per step in each of the four axis directions (Fig.~\ref{fig:setup}b). During each time-step, FLoWM's memory does four operations: \emph{Move}:~every copy shifts by its own velocity, and all copies shift or rotate against the agent's motion. These built-in shifts carry a block's content while it is out of view. \emph{Mix}:~a learned $3\times3$ convolution combines the copies, so content can change copies when a block bounces. \emph{Render}:~in the prediction phase, a learned decoder renders the predicted frame from the cells in view. \emph{Write}:~a learned transformer writes a frame into the cells in view. It writes the camera frame during context and its own rendered frame during prediction, which we call the \emph{write-back}.

The memory never stores a position as a number, so a probe must read it. The memory can carry only motion that the five copies and the mix can express. Its size grows with the area of the room times the number of velocities, not with the number of objects.

\subsection{A node memory with oracle perception}
\label{sec:node}

A node memory stores one node per object (Fig.~\ref{fig:setup}c). Each node holds a position and a velocity as numbers, their uncertainties, and appearance features. A node moves by its own velocity, so no velocity is built into the memory's layout. The tracker keeps one Kalman filter~\citep{kalman1960} per block as its node, called a track. It matches detections to tracks by distance and color, in the style of SORT~\citep{bewley2016sort}. It has no trained networks. The simulator provides what a learned node memory must produce. These oracles are block masks, positions with one pixel of noise and their uncertainty, the walls, and at readout each detection's identity, linking tracks to blocks (App.~\ref{app:tracker}). A minimal working node memory therefore still needs four learned parts: a detector that finds blocks and estimates their floor positions from pixels, a readout that links nodes to objects without true identities, a bounce model that does not know the walls, and velocity estimates that hold up without hand-set allowed speeds.

Like FloWM, the tracker has a prior over speeds. A small error in a velocity estimated from noisy positions grows into a large position error over a long gap without sightings. The tracker therefore snaps each axis of a track's velocity to an allowed speed, which we call \emph{speed snapping}. On the standard setting the allowed speeds are 0 and 1 cell per step, fitted on training videos.

\section{Results}
\label{sec:results}

\begin{figure}[t]
\begin{minipage}[t]{0.565\textwidth}
\vspace{0pt}
\centering
\footnotesize
\captionof{table}{Accuracy (\%). Top: standard test videos and state size. Bottom: varied test videos, out of view. $^a$Convolutional probe trained on the standard setting. $^b$Probes trained on varied videos.}
\label{tab:main}
\vspace{4pt}
\setlength{\tabcolsep}{3.1pt}
\begin{tabular}{@{}lccc@{}}
\toprule
\emph{Standard setting} & In view & Out of view & State \\
\midrule
FloWM, convolutional probe & 96.5 & 88.6 & 5.6MB \\
FloWM, single-cell probe & 95.7 & 66.1 & 5.6MB\\
Tracker & 90.9 & 90.8 & 14KB \\
\midrule
\emph{Varied setting, out of view} & Built-in & Diagonal & Speed 2 \\
\midrule
FloWM, released$^a$ & 72.7 & 3.5 & 4.2 \\
FloWM, fine-tuned$^b$ & 39.1 & 11.0 & 13.0 \\
FloWM, retrained$^b$ & 16.8 & 8.2 & 9.4 \\
Tracker, speeds 0 and 1 & 89.9 & 93.1 & 4.9 \\
Tracker, speeds 0, 1 and 2 & 91.0 & 94.1 & 93.9 \\
\bottomrule
\end{tabular}
\end{minipage}\hfill
\begin{minipage}[t]{0.405\textwidth}
\vspace{0pt}
\centering
\includegraphics[width=\linewidth]{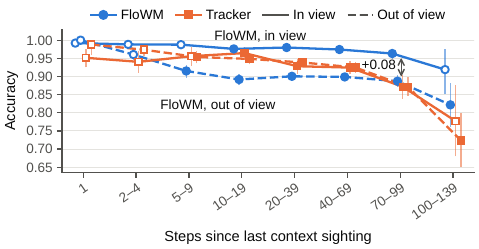}
\captionof{figure}{Accuracy by steps since a block's last sighting, on the standard test videos, with FloWM read by the convolutional probe. Solid lines are blocks in view and dashed lines blocks out of view. The tracker's two lines stay close from 5 to 99 steps, and the annotation marks FloWM's in-view lead. Bars are 95\% intervals, and hollow markers mark sparse bins.}
\label{fig:recency}
\end{minipage}
\end{figure}

\subsection{Does the grid hold a position, or does the probe compute it?}
\label{sec:q1}

We used two probes to answer this question. A \emph{convolutional probe}, rebuilt from FloWM's appendix, sees the whole grid and has 7.06M parameters per head, more than half of FloWM's own 12.8M. A \emph{single-cell probe} applies one linear map of 1,282 parameters to each cell's own features. It finds a block only where the memory stores it in the block's own cell. We trained one probe head per block color on frozen memory states (App.~\ref{app:probes}). For blocks in view the single-cell probe finds the position, and for blocks out of view the convolutional probe computes part of it. The two probes agree on blocks in view and disagree out of view (Table~\ref{tab:main}, top). There the single-cell probe finds 66\% of blocks and the convolutional probe 89\%. The extra hits need neighboring cells or a nonlinear combination of features.

No block is observed during prediction, yet FloWM, read by the convolutional probe, scores 8 points higher in view than out of view. The difference holds at every time since the last sighting from 5 steps on (Fig.~\ref{fig:recency})\posthoc, while the tracker shows none. FloWM's score also rises by 5 points when a block enters view\posthoc (App.~\ref{app:writeback}). We attribute both to the write-back, which rewrites the cells in view from FloWM's own rendered frames. The write-back adds no observation, but in-view scores depend on rendering as well as memory, so we compare memories out of view for fairness. {\let\thefootnote\relax\footnotetext{$\dagger$ marks a choice made after seeing results (App.~\ref{app:posthoc})}}

\subsection{On FloWM's own setting, how close does a node memory come?}
\label{sec:q2}

With oracle perception, a node memory reaches FloWM's range. Out of view the tracker finds 91\% of blocks, and FloWM, read by the convolutional probe, finds 89\% (Table~\ref{tab:main}, top). We call this the same range and not a win, because our rebuilt probe scores 90\% overall where FloWM's authors report 96\%. In view FloWM leads by about 6 points. The inputs also differ, since FloWM perceives from pixels and the tracker receives oracle detections. The tracker's state takes 14KB against 5.6MB for FloWM's five copies of the grid (App.~\ref{app:cost}), from which FloWM also renders frames. The tracker's positions are numbers, so no probe is needed to read them once tracks are linked to objects.

\subsection{What happens outside the built-in velocities?}
\label{sec:q3}

Each memory loses blocks that move outside its motion prior, and the known fixes differ in cost. On varied test videos, FloWM's released model finds 73\% of built-in blocks but 4\% of diagonal and speed-2 blocks (Table~\ref{tab:main}, bottom). Using MSE, its prediction error on frames is 23 times higher than on the standard setting. Fine-tuning or retraining on varied videos lowers it, yet probes trained on those videos find at most 13\% of diagonal and speed-2 blocks.

The tracker loses speed-2 blocks in the same way. Unchanged, with allowed speeds 0 and 1, it keeps diagonal blocks, which per-axis speeds can express, and finds 5\% of speed-2 blocks. Without speed snapping it finds about 60\% of them, so for the tracker a wrong prior costs more than none (App.~\ref{app:varied}). With the allowed speeds refitted to 0, 1 and 2 on varied training videos, it finds 94\% of diagonal and speed-2 blocks. With exact positions the tracker needs no allowed speeds and finds 97\% of all blocks. Covering the varied setting's 24 velocities with built-in shifts would take 25 copies of the grid, instead of five~\citep[their Apps.~F and G.1]{lillemark2026flowm}. Training that model is out of scope for this paper.

\subsection{What stands between the tracker and the full platform?}
\label{sec:q4}

A full node-based memory platform must replace every oracle in Sec.~\ref{sec:node}, and solve three more problems. The first is velocity under noise. Without speed snapping the tracker finds 47\% of out-of-view blocks on the standard setting, not 91\% (App.~\ref{app:tracker}), and snapping assumes few, discrete speeds.

The second is long gaps. Out of view, the tracker leads FloWM by 6 points at 10 to 19 steps since the last sighting and trails by 10 points after 100\posthoc{} (Fig.~\ref{fig:recency}). Two causes explain the late deficit\posthoc. One sighting gives the tracker a position but no velocity. At long gaps FloWM finds 30\% of blocks sighted once, the tracker almost none. Tuning videos suggest FloWM infers their direction (App.~\ref{app:recency}). Other tracks pick up error after the last sighting, as when another block's detection joins them.

The third is the bounce. We replaced the exact reflection off the true walls with a bounce trained on the track's position error at each new sighting\posthoc. We ran this on training videos with up to ten blocks and repeated colors. The share of matches that join a detection to another block's track rises from 16\% to between 19\% and 26\%, against 17.5\% without a bounce model (App.~\ref{app:bounce}).

\section{Conclusion}
\label{sec:conclusion}

On the standard setting, where both priors cover the motion, a node memory with oracle perception scores in the same range as FloWM, which works from pixels. When the prior is wrong, both memories struggle to accurately track blocks. Our evidence qualifies the workshop's Motion~2: explicit dynamics generalize only as far as their motion prior reaches for both memories. We tested new dynamics, not new goals, and no implicit model. It supports Motion~6: the same memory scores differently with a different probe, and FloWM's in-view scores rise through its own write-back, not through memory. Fair comparisons should report the probe and score objects out of view.

\textbf{Limitations.}
We used one synthetic environment and one released FloWM model. The tracker's inputs are oracles (Sec.~\ref{sec:node}). The two probes differ in both reach and capacity, so the single-cell probe isolates neither. We froze the tracker's settings for the standard setting after seeing some test results. App.~\ref{app:posthoc} lists the choices marked $\dagger$ and gives the intervals we have.

\label{bodyend}


\label{refstart}
\bibliography{refs}

@inproceedings{lillemark2026flowm,
  title     = {Flow Equivariant World Models: Memory for Partially Observed Dynamic Environments},
  author    = {Lillemark, Hansen Jin and Huang, Benhao and Zhan, Fangneng and Du, Yilun and Keller, Thomas Anderson},
  booktitle = {International Conference on Machine Learning (ICML)},
  year      = {2026},
  eprint    = {2601.01075},
  archivePrefix = {arXiv},
  primaryClass  = {cs.LG}
}

@inproceedings{chevalierboisvert2023minigrid,
  title     = {{Minigrid} \& {Miniworld}: Modular \& Customizable Reinforcement Learning Environments for Goal-Oriented Tasks},
  author    = {Chevalier-Boisvert, Maxime and Dai, Bolun and Towers, Mark and de Lazcano, Rodrigo and Willems, Lucas and Lahlou, Salem and Pal, Suman and Castro, Pablo Samuel and Terry, Jordan},
  booktitle = {Advances in Neural Information Processing Systems (NeurIPS), Datasets and Benchmarks Track},
  year      = {2023},
  eprint    = {2306.13831},
  archivePrefix = {arXiv}
}

@misc{duan2026liveworld,
  title     = {{LiveWorld}: Simulating Out-of-Sight Dynamics in Generative Video World Models},
  author    = {Duan, Zicheng and Xia, Jiatong and Zhang, Zeyu and Zhang, Wenbo and Zhou, Gengze and Gou, Chenhui and He, Yefei and Chen, Feng and Zhang, Xinyu and Liu, Lingqiao},
  year      = {2026},
  eprint    = {2603.07145},
  howpublished = {arXiv:2603.07145},
  archivePrefix = {arXiv},
  primaryClass  = {cs.CV}
}

@misc{creswell2021oat,
  title     = {Unsupervised Object-Based Transition Models for {3D} Partially Observable Environments},
  author    = {Creswell, Antonia and Kabra, Rishabh and Burgess, Chris and Shanahan, Murray},
  year      = {2021},
  eprint    = {2103.04693},
  howpublished = {arXiv:2103.04693},
  archivePrefix = {arXiv},
  primaryClass  = {cs.LG}
}

@inproceedings{daniel2026lpwm,
  title     = {Latent Particle World Models: Self-supervised Object-centric Stochastic Dynamics Modeling},
  author    = {Daniel, Tal and Qi, Carl and Haramati, Dan and Zadeh, Amir and Li, Chuan and Tamar, Aviv and Pathak, Deepak and Held, David},
  booktitle = {International Conference on Learning Representations (ICLR)},
  year      = {2026},
  eprint    = {2603.04553},
  archivePrefix = {arXiv}
}

@misc{chen2026hydra,
  title     = {Out of Sight but Not Out of Mind: Hybrid Memory for Dynamic Video World Models},
  author    = {Chen, Kaijin and Liang, Dingkang and Zhou, Xin and Ding, Yikang and Liu, Xiaoqiang and Wan, Pengfei and Bai, Xiang},
  year      = {2026},
  eprint    = {2603.25716},
  howpublished = {arXiv:2603.25716},
  archivePrefix = {arXiv},
  primaryClass  = {cs.CV}
}

@inproceedings{nishimoto2026stica,
  title     = {Object-Centric World Models for Causality-Aware Reinforcement Learning},
  author    = {Nishimoto, Yosuke and Matsubara, Takashi},
  booktitle = {Proceedings of the AAAI Conference on Artificial Intelligence (AAAI)},
  year      = {2026},
  eprint    = {2511.14262},
  archivePrefix = {arXiv}
}

@inproceedings{xiao2025worldmem,
  title     = {{WorldMem}: Long-term Consistent World Simulation with Memory},
  author    = {Xiao, Zeqi and Lan, Yushi and Zhou, Yifan and Ouyang, Wenqi and Yang, Shuai and Zeng, Yanhong and Pan, Xingang},
  booktitle = {Advances in Neural Information Processing Systems (NeurIPS)},
  year      = {2025},
  eprint    = {2504.12369},
  archivePrefix = {arXiv}
}

@misc{huang2025memoryforcing,
  title     = {Memory Forcing: Spatio-Temporal Memory for Consistent Scene Generation on {Minecraft}},
  author    = {Huang, Junchao and Hu, Xinting and Han, Boyao and Shi, Shaoshuai and Tian, Zhuotao and He, Tianyu and Jiang, Li},
  year      = {2025},
  eprint    = {2510.03198},
  howpublished = {arXiv:2510.03198},
  archivePrefix = {arXiv},
  primaryClass  = {cs.CV}
}

@misc{lu2026wrbench,
  title     = {Current World Models Lack a Persistent State Core},
  author    = {Lu, Jinpeng and Zhu, Dexu and Shi, Haoyuan and Cai, Linghan and Tang, Guo and Chen, Yinda and Cao, Jie and Tang, Duyu and Zhang, Yi and Dai, Yong and Ju, Xiaozhu},
  year      = {2026},
  eprint    = {2606.20545},
  howpublished = {arXiv:2606.20545},
  archivePrefix = {arXiv}
}

@misc{ma2026stevo,
  title     = {Out of Sight, Out of Mind? Evaluating State Evolution in Video World Models},
  author    = {Ma, Ziqi and Liufu, Mengzhan and Gkioxari, Georgia},
  year      = {2026},
  eprint    = {2603.13215},
  howpublished = {arXiv:2603.13215},
  archivePrefix = {arXiv},
  primaryClass  = {cs.CV}
}

@misc{chen2026memobench,
  title     = {{MemoBench}: Benchmarking World Modeling in Dynamically Changing Environments},
  author    = {Chen, Haoyu and Zhou, Kaichen and Hua, Hang and Zhang, Kaile and Qian, Jingwen and Ma, Wufei and Chen, Haonan and Liu, Chunjiang and Zhao, Yizhou and Wang, Xiaoyuan and Li, Weiyue and Yuille, Alan and Liang, Paul Pu and Du, Yilun},
  year      = {2026},
  eprint    = {2606.27537},
  howpublished = {arXiv:2606.27537},
  archivePrefix = {arXiv}
}

@misc{gu2026r2mbench,
  title     = {{R2M-Bench}: Evaluating Revisit Memory via Relative Consistency in Interactive Video World Models},
  author    = {Gu, Qiwen and Gao, Bingjie and Chen, Rui and Li, Geng and Li, Jifan and Wen, Qishuai and Niu, Li and Tang, Jing and Chu, Xiangxiang and Zhao, Junqiao},
  year      = {2026},
  eprint    = {2608.27328},
  howpublished = {arXiv:2608.27328},
  archivePrefix = {arXiv}
}

@misc{lian2025loopnav,
  title     = {{LoopNav}: Benchmarking Spatial Consistency in World Models},
  author    = {Lian, Kewei and Cai, Shaofei and Liang, Yitao and Liu, Anji},
  year      = {2025},
  eprint    = {2505.22976},
  howpublished = {arXiv:2505.22976},
  archivePrefix = {arXiv}
}

@misc{king2026echomemory,
  title     = {{Echo-Memory}: A Controlled Study of Memory in Action World Models},
  author    = {King, Wayne and Xue, Zeyue and Bian, Yuxuan and Huang, Jie and Li, Haoran and Li, Yaowei and Su, Yaofeng and Li, Yuming and Wang, Haoyu and Zhang, Shiyi and Zhang, Songchun and Niu, Yuwei and Xu, Sihan and Zhuang, Junhao and Huang, Haoyang and Duan, Nan},
  year      = {2026},
  eprint    = {2606.09803},
  howpublished = {arXiv:2606.09803},
  archivePrefix = {arXiv}
}

@inproceedings{vafa2024evaluating,
  title     = {Evaluating the World Model Implicit in a Generative Model},
  author    = {Vafa, Keyon and Chen, Justin Y. and Rambachan, Ashesh and Kleinberg, Jon and Mullainathan, Sendhil},
  booktitle = {Advances in Neural Information Processing Systems (NeurIPS)},
  volume    = {37},
  year      = {2024},
  eprint    = {2406.03689},
  archivePrefix = {arXiv}
}

@inproceedings{zhang2026worldinworld,
  title     = {World-in-World: World Models in a Closed-Loop World},
  author    = {Zhang, Jiahan and Jiang, Muqing and Dai, Nanru and Lu, Taiming and Uzunoglu, Arda and Zhang, Shunchi and Wei, Yana and Wang, Jiahao and Patel, Vishal M. and Liang, Paul Pu and Khashabi, Daniel and Peng, Cheng and Chellappa, Rama and Shu, Tianmin and Yuille, Alan and Du, Yilun and Chen, Jieneng},
  booktitle = {International Conference on Learning Representations (ICLR)},
  year      = {2026},
  eprint    = {2510.18135},
  archivePrefix = {arXiv}
}

@inproceedings{hewitt2019control,
  title     = {Designing and Interpreting Probes with Control Tasks},
  author    = {Hewitt, John and Liang, Percy},
  booktitle = {Proceedings of the 2019 Conference on Empirical Methods in Natural Language Processing and the 9th International Joint Conference on Natural Language Processing (EMNLP-IJCNLP)},
  pages     = {2733--2743},
  year      = {2019},
  doi       = {10.18653/v1/D19-1275}
}

@article{belinkov2022probing,
  title     = {Probing Classifiers: Promises, Shortcomings, and Advances},
  author    = {Belinkov, Yonatan},
  journal   = {Computational Linguistics},
  volume    = {48},
  number    = {1},
  pages     = {207--219},
  year      = {2022},
  doi       = {10.1162/coli_a_00422}
}

@inproceedings{li2023othello,
  title     = {Emergent World Representations: Exploring a Sequence Model Trained on a Synthetic Task},
  author    = {Li, Kenneth and Hopkins, Aspen K. and Bau, David and Vi{\'e}gas, Fernanda and Pfister, Hanspeter and Wattenberg, Martin},
  booktitle = {International Conference on Learning Representations (ICLR)},
  year      = {2023},
  eprint    = {2210.13382},
  archivePrefix = {arXiv}
}

@inproceedings{bewley2016sort,
  title     = {Simple Online and Realtime Tracking},
  author    = {Bewley, Alex and Ge, Zongyuan and Ott, Lionel and Ramos, Fabio and Upcroft, Ben},
  booktitle = {IEEE International Conference on Image Processing (ICIP)},
  year      = {2016},
  eprint    = {1602.00763},
  archivePrefix = {arXiv}
}

@inproceedings{ristani2016idf1,
  title     = {Performance Measures and a Data Set for Multi-Target, Multi-Camera Tracking},
  author    = {Ristani, Ergys and Solera, Francesco and Zou, Roger S. and Cucchiara, Rita and Tomasi, Carlo},
  booktitle = {European Conference on Computer Vision (ECCV) Workshops},
  year      = {2016},
  eprint    = {1609.01775},
  archivePrefix = {arXiv}
}

@inproceedings{locatello2020slot,
  title     = {Object-Centric Learning with Slot Attention},
  author    = {Locatello, Francesco and Weissenborn, Dirk and Unterthiner, Thomas and Mahendran, Aravindh and Heigold, Georg and Uszkoreit, Jakob and Dosovitskiy, Alexey and Kipf, Thomas},
  booktitle = {Advances in Neural Information Processing Systems (NeurIPS)},
  year      = {2020},
  eprint    = {2006.15055},
  archivePrefix = {arXiv}
}

@inproceedings{kipf2020cswm,
  title     = {Contrastive Learning of Structured World Models},
  author    = {Kipf, Thomas and van der Pol, Elise and Welling, Max},
  booktitle = {International Conference on Learning Representations (ICLR)},
  year      = {2020},
  eprint    = {1911.12247},
  archivePrefix = {arXiv}
}

@article{hafner2025dreamerv3,
  title     = {Mastering Diverse Control Tasks through World Models},
  author    = {Hafner, Danijar and Pasukonis, Jurgis and Ba, Jimmy and Lillicrap, Timothy},
  journal   = {Nature},
  volume    = {640},
  pages     = {647--653},
  year      = {2025},
  doi       = {10.1038/s41586-025-08744-2}
}

@inproceedings{zhou2025dinowm,
  title     = {{DINO-WM}: World Models on Pre-trained Visual Features enable Zero-shot Planning},
  author    = {Zhou, Gaoyue and Pan, Hengkai and LeCun, Yann and Pinto, Lerrel},
  booktitle = {International Conference on Machine Learning (ICML)},
  year      = {2025},
  eprint    = {2411.04983},
  archivePrefix = {arXiv}
}

@misc{assran2025vjepa2,
  title     = {{V-JEPA} 2: Self-Supervised Video Models Enable Understanding, Prediction and Planning},
  author    = {Assran, Mido and Bardes, Adrien and Fan, David and Garrido, Quentin and Howes, Russell and Komeili, Mojtaba and Muckley, Matthew and Rizvi, Ammar and Roberts, Claire and Sinha, Koustuv and Zholus, Artem and Arnaud, Sergio and Gejji, Abha and Martin, Ada and Hogan, Francois Robert and Dugas, Daniel and Bojanowski, Piotr and Khalidov, Vasil and Labatut, Patrick and Massa, Francisco and Szafraniec, Marc and Krishnakumar, Kapil and Li, Yong and Ma, Xiaodong and Chandar, Sarath and Meier, Franziska and LeCun, Yann and Rabbat, Michael and Ballas, Nicolas},
  year      = {2025},
  eprint    = {2506.09985},
  howpublished = {arXiv:2506.09985},
  archivePrefix = {arXiv}
}

@inproceedings{gregor2019shaping,
  title     = {Shaping Belief States with Generative Environment Models for {RL}},
  author    = {Gregor, Karol and Rezende, Danilo Jimenez and Besse, Frederic and Wu, Yan and Merzic, Hamza and van den Oord, A{\"a}ron},
  booktitle = {Advances in Neural Information Processing Systems (NeurIPS)},
  year      = {2019},
  eprint    = {1906.09237},
  archivePrefix = {arXiv}
}

@inproceedings{wijmans2023emergence,
  title     = {Emergence of Maps in the Memories of Blind Navigation Agents},
  author    = {Wijmans, Erik and Savva, Manolis and Essa, Irfan and Lee, Stefan and Morcos, Ari S. and Batra, Dhruv},
  booktitle = {International Conference on Learning Representations (ICLR)},
  year      = {2023},
  eprint    = {2301.13261},
  archivePrefix = {arXiv}
}

@inproceedings{shamsian2020permanence,
  title     = {Learning Object Permanence from Video},
  author    = {Shamsian, Aviv and Kleinfeld, Ofri and Globerson, Amir and Chechik, Gal},
  booktitle = {European Conference on Computer Vision (ECCV)},
  year      = {2020}
}

@inproceedings{tokmakov2021permatrack,
  title     = {Learning to Track with Object Permanence},
  author    = {Tokmakov, Pavel and Li, Jie and Burgard, Wolfram and Gaidon, Adrien},
  booktitle = {IEEE/CVF International Conference on Computer Vision (ICCV)},
  year      = {2021}
}

@article{kalman1960,
  author  = {Kalman, Rudolf E.},
  title   = {A New Approach to Linear Filtering and Prediction Problems},
  journal = {Journal of Basic Engineering},
  volume  = {82},
  number  = {1},
  pages   = {35--45},
  year    = {1960}
}

\clearpage
\appendix

\section{Environment, components and protocol}
\label{app:protocol}

\subsection{Environment}
Dynamic Block World~\citep{lillemark2026flowm} is a room built in Miniworld~\citep{chevalierboisvert2023minigrid} (Fig.~\ref{fig:setup}a). The floor is a grid of cells, and the walls are 16 cells apart. Colored blocks, each 0.7 cells wide, slide across the floor and bounce off the walls. An agent moves through the room in steps of one whole cell and turns in steps of $90^\circ$. It sees the room through a camera with a $60^\circ$ field of view, as $128\times128$ color frames.

Each video has two phases. In the \emph{context phase}, a model sees 70 frames together with the agent's actions. In the \emph{prediction phase}, the model receives only the agent's next 70 actions and sees no frames.  We use two settings that differ only in how the blocks move. 

In the \emph{standard setting}, which follows the generator settings of FloWM's published data, every block moves one cell per step along one of the two floor axes. We created a \emph{varied setting}, in which each block gets a random nonzero velocity whose components are whole numbers of cells per step between $-2$ and $2$. Only 16.9\% of blocks in the varied setting move as they would in the standard setting. Another 15.8\% move diagonally, one cell per step on each axis. The remaining 67.3\% move two cells per step on at least one axis. We generated both settings with FloWM's own version of Miniworld. In the varied setting, blocks faster than one cell per step move in one-cell sub-steps, so that they bounce off the walls instead of passing through them.

\subsection{FloWM's memory in detail}
FloWM keeps its memory as a top-down map of the floor around the agent. The map is a $33\times33$ grid of cells centered on the agent, and one cell of the map covers one cell of the floor. Each cell holds a vector of 256 learned features, and content about a block is meant to sit in the cell where the block is. The memory holds five copies of this grid. Each copy is tied to one velocity from a fixed set, the \emph{built-in velocities}.
The five velocities are standing still and moving one cell per step in each of the four axis directions (Fig.~\ref{fig:setup}b). A block that moves in the $+x$ direction should therefore be stored in the $+x$ copy.

Every step, FloWM first moves and mixes the copies. In the context phase, it then writes the camera frame into the memory. In the prediction phase, it then renders a frame from the memory and writes that frame back into the memory.

\begin{itemize}
    \item \emph{Move.} FloWM shifts each copy by one cell in that copy's direction and leaves the still copy in place. It also moves all copies against the agent's own motion, so that the grid stays centered on the agent: a step forward shifts the grid by one cell, and a turn rotates the grid by $90^\circ$ and turns each copy's direction with it. These moves are built in and are not learned. They carry a block's content along with the block, even while the block is out of view.
    \item \emph{Mix.} A learned $3\times3$ convolution combines the five copies, separately for each of the 256 features (57.6k parameters). This step can move content from one copy to another, which a bounce requires, since a block that hits a wall must move from the $+x$ copy to the $-x$ copy. Because the convolution spans neighboring cells, it can also move content by up to one cell. The mix applies the same weights at every cell, however, so it cannot tell where a wall is, and the content that turns a block at a wall must come from what the write stored.
    \item \emph{Render.} During the prediction phase, a learned decoder renders the predicted camera frame from the cells in view, after the move and the mix and before that step's write.
    \item \emph{Write.} A learned transformer writes a frame into the memory. It updates only a fixed set of 177 cells, a $60^\circ$ wedge in front of the agent that covers the camera's view, in all five copies. During the context phase it writes the real camera frame. During the prediction phase there is no camera frame, so FloWM writes the frame it has just rendered at the same step. We call this the \emph{write-back}.
\end{itemize}

FloWM is trained end to end to predict frames, with the mean squared error (MSE) between predicted and true frames as its loss. It has 12.8M parameters, and its memory holds $33\times33\times5\times256\approx1.4$M numbers. The velocity copies matter, since removing them raises the prediction error 12.6-fold~\citep[Table~2]{lillemark2026flowm}. We use the released checkpoint, which reproduces the published MSE of 0.000603 on the release's own validation clips.

Three properties of this memory matter for our study. First, it never stores a block's position as a number. The position is implied by which cells hold the block's content, so a probe is needed to read it. Second, the memory can only carry motion that its five copies, together with the mix, can express. Third, its size grows with the area of the room and the number of velocities, not with the number of objects.

\subsection{Where each part of the two memories comes from}
Table~\ref{tab:components} lists which parts of the tracker the simulator provides (oracle), which we wrote by hand, and which we fitted on oracle data, and it compares each part with FloWM.

\begin{table}[h]
    \centering
    \footnotesize
    \caption{Where each part of the two memories comes from. \emph{Oracle} parts are given by the simulator. \emph{Hand-set} parts are written by us, with no training. \emph{Fitted} parts are estimated from noisy oracle positions on training videos of the standard setting, with thresholds chosen on the tuning videos. \emph{Built in} parts are fixed by FloWM's design. \emph{Learned} parts are trained with FloWM's frame-prediction loss, except the probe, which is trained to read positions. The table describes the tracker with allowed speeds 0 and 1.}
    \label{tab:components}
        \begin{tabular}{>{\raggedright\arraybackslash}p{0.27\linewidth}>{\raggedright\arraybackslash}p{0.33\linewidth}>{\raggedright\arraybackslash}p{0.30\linewidth}}
        \toprule
        Part & Tracker & FloWM \\
        \midrule
        Finding blocks in a frame & Oracle: true masks & Learned: write transformer \\
        Block position in a frame & Oracle: true position plus one-pixel noise & Learned: write transformer \\
        Block appearance & Oracle: mean color inside the true mask & Learned: 256 features per cell \\
        Agent's own motion & Hand-set: from actions & Built in: shift and rotation by actions \\
        Motion between steps & Hand-set: Kalman predict step & Built in: five velocity copies \\
        Allowed speeds & Fitted: snap each axis to 0 or 1 & Built in: the five velocities \\
        Bounces & Oracle: true walls, exact reflection & Learned: write transformer and mix layer \\
        Linking a sighting to memory & Hand-set: SORT-style matching & Learned: write transformer \\
        Reading out a position & Oracle: track last matched by the block & Learned: probe per color \\
        \bottomrule
    \end{tabular}
\end{table}

\subsection{Task and scoring}
\begin{figure}[h]
\centering
\begin{tikzpicture}[x=0.82mm,y=1cm,font=\footnotesize]
  \fill[cB!12] (0,0) rectangle (70,0.55);
  \fill[cK!18] (70,0) rectangle (140,0.55);
  \draw (0,0) rectangle (140,0.55);
  \draw (70,0) -- (70,0.55);
  \node at (35,0.275) {Context: 70 frames and actions};
  \node at (105,0.275) {Prediction: 70 actions, no frames};
  \node[below,font=\scriptsize] at (0,0) {step 1};
  \node[below,font=\scriptsize] at (70,0) {70};
  \node[below,font=\scriptsize] at (140,0) {140};
  \foreach \s in {6,7,22,23,24,41} \fill[cR] (\s,0.78) circle (1.5pt);
  \node[anchor=south west,cR,font=\scriptsize] at (0,0.86) {sightings of one block};
  \draw[cR,densely dashed] (41,0.78) -- (41,0.55) (41,0) -- (41,-0.45);
  \draw[cP,thick] (100,0.78) -- (100,0.55) (100,0) -- (100,-0.45);
  \fill[cP] (100,0.78) circle (1.5pt);
  \node[anchor=south,cP,font=\scriptsize,align=center] at (100,0.82) {scored step: is the named cell\\the block's true cell?};
  \draw[decorate,decoration={brace,mirror,amplitude=4pt}] (41,-0.5) -- (100,-0.5)
    node[midway,below=5pt,font=\scriptsize] {steps since last sighting $=59$};
\end{tikzpicture}
\caption{The protocol for one block. Both memories receive the same context phase. FloWM receives the frames, and the tracker receives oracle detections from the same frames. At every predicted step, each memory names one grid cell for the block, and the answer is correct only if it is the block's exact cell. A sighting is a context frame in which the block covers at least 12 pixels. We split answers by whether the block lies inside the camera's view at the scored step and by the steps since the block's last sighting.}
\label{fig:timeline}
\end{figure}
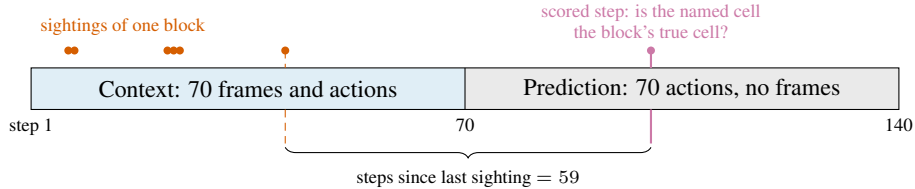

Figure~\ref{fig:timeline} shows the protocol. Both memories receive the same videos. During the context phase, FloWM receives the 70 frames and actions, and the tracker receives oracle detections from the same 70 frames together with the same actions. During the prediction phase, both receive only the next 70 actions.

At every one of the 70 predicted steps, each memory names one cell of the agent-centered $33\times33$ grid for each block, or states that the block is off the grid. This gives 1,090 possible answers. An answer is correct only if it is the cell that contains the block's center at that step, counted in cells forward and to the right of the agent, so an answer one cell away counts as wrong.
The room is smaller than the grid, so the correct answer is never off the grid. Accuracy is the fraction of correct answers over all blocks and all predicted steps. Guessing at random would score below 0.1\%.

Each video that we score with probes or the tracker contains six blocks, one of each color (red, blue, green, purple, gray and yellow), so each block can be identified by its color. We use 978 videos from the standard setting as a \emph{tuning set} for training probes and for choosing the tracker's settings. The tracker's allowed speeds and the learned bounce use separate training videos (Apps.~\ref{app:tracker} and~\ref{app:bounce}). We report results on 256 fresh videos from the standard setting, which we call the \emph{test set}, and on 1,024 fresh videos from the varied setting. Confidence intervals are 95\% intervals from resampling whole videos (a video bootstrap).

\subsection{Splitting the score}
We split scores on the standard setting in three ways.

\emph{In view or out of view.} A block is in view at a predicted step if it lies inside the camera's $60^\circ$ field of view at that step, even if another block hides it. During the prediction phase no frames are shown, so a block in view is no more observed than a block behind the agent. About four in five answers on the test set are out of view (85,602 of 107,520).

\emph{Steps since the last sighting.} A sighting is a context frame in which the block's true mask covers at least 12 pixels. For each answer, we count the steps from the block's last sighting to the scored step (Fig.~\ref{fig:timeline}). On the test set this count runs from 1 to 137. Sixteen of the 1,536 test blocks are never sighted during the context phase, and they give about 1\% of answers, both in view and out of view. These answers have no count, so they appear in neither this split nor the split by number of sightings, but they stay in every other score. The tracker has no track for such a block, so all its answers for it are wrong, while FloWM finds 5.0\% of them in view and 2.2\% out of view.

\section{Probes}
\label{app:probes}

FloWM's memory holds features rather than positions, so we need a probe to read a position out of it. We train one probe head per block color. Each head looks at the frozen memory at a predicted step, after that step's write-back, and outputs one of the 1,090 answers for its color's block. Heads are trained on memory states saved from the tuning videos and are then scored, unchanged, on the test videos. We use two probes.

The \emph{convolutional probe} is the probe described in FloWM's appendix. It is a convolutional network that sees the whole grid, with 7.06M parameters per head, which is more than half the size of FloWM itself (12.8M). Its code was not released, so we rebuilt it. Our rebuild scores 90.2\% on the test set, while FloWM's paper reports 96\%.

The \emph{single-cell probe} scores each cell using only that cell's own features in the five copies. It applies one small linear map, shared by all cells, and has 1,282 parameters per head. It cannot gather evidence from neighboring cells, and it cannot combine features in a nonlinear way. It can therefore find a block only where the memory stores that block in the correct cell.

The tracker needs no probe. Its answer for a block is the grid cell that contains the position of the block's last matched track, measured from the agent's position and heading as computed from its actions, and rounded as the labels are. A block that was never sighted has no track, so its answers count as wrong.

\subsection{The probe changes the score}
The two probes agree on blocks in view, where the single-cell probe scores 95.7\% and the convolutional probe scores 96.5\% (Table~\ref{tab:main}). They disagree on blocks out of view. The single-cell probe finds 66.1\% of out-of-view blocks, while the convolutional probe finds 88.6\%. About a quarter of the convolutional probe's out-of-view hits therefore need information from neighboring cells or a nonlinear combination of features. For these blocks, the memory does not store the block in its own cell in a form that a simple read can find, and the probe computes part of the answer.

This matters when we compare memories. Out of view, the tracker scores 90.8\%. It is 2.1 points above FloWM read by the convolutional probe (CI 0.5 to 3.7), and about 25 points above FloWM read by the single-cell probe. We do not read the first gap as a win for the tracker, because our rebuilt probe scores below the published one. Which memory looks better therefore depends on the probe. Counting all answers in view and out of view, the convolutional probe scores 90.2\%, the single-cell probe 72.1\% and the tracker 90.8\%.

\section{The tracker with oracle perception}
\label{app:tracker}

A node memory takes the opposite approach to FloWM's grid.
It stores one node per object. Each node holds the object's position and velocity as explicit numbers, an uncertainty for each, and features that describe the object's appearance. Each node moves by its own velocity, so the memory itself does not limit which velocities it can carry, and its size grows with the number of objects rather than the area of the room. A learned node memory must solve several core problems. It must find objects in the image, estimate their floor positions from the image, learn how they move and bounce, and decide which node each new sighting belongs to.

To separate the bookkeeping from these perception problems, we built the simplest node memory and gave it perception for free. This is a Kalman tracker that receives true block positions and masks from the simulator. It contains no trained networks.

\subsection{The Kalman filter}
A Kalman filter~\citep{kalman1960} keeps a best guess of an object's state together with a measure of how uncertain that guess is. In the tracker, the state of each block is its floor position and velocity, $(x, y, v_x, v_y)$, and the uncertainty is a $4\times4$ covariance matrix. The filter alternates two steps. The \emph{predict} step moves the guess forward by one time step using the current velocity, and it increases the uncertainty to allow for motion that the filter cannot foresee. The \emph{update} step receives a measured position and moves the guess toward it. The size of this move depends on the two uncertainties. A confident prediction moves only a little, while a confident measurement moves it a lot. The filter never measures velocity directly. It infers velocity from how the measured position changes across sightings. A track that receives only one detection therefore keeps its starting velocity of zero, and it stays where the block was seen.

\subsection{Matching and tracks}
A tracker runs one filter per object and must decide which detection belongs to which filter. Our matching is in the style of SORT~\citep{bewley2016sort}, and the tracker repeats four steps at every time step. First, it predicts every track forward, and when a track's center would cross a line one cell inside a wall, where a real block's center turns around, it mirrors the track back and reverses its velocity on that axis. Second, it matches the detections in the current frame to tracks. The cost of a pairing grows with the distance between the detection and the track's predicted position, measured against their combined uncertainty, and with their difference in mean color, which counts more the longer the track has gone unseen. Pairs farther apart than the 99\% range of that uncertainty are not allowed, and the tracker repeatedly takes the cheapest allowed pair, so it assigns at most one detection to each track. A leftover detection may still join a track that has gone unseen for several steps, under a wider distance limit, if their colors nearly match. Third, it updates each matched track with its detection. Fourth, it starts a new track for each detection that matched no track. A track whose block is out of view receives no detection of that block, so it only predicts, unless a detection of another block is matched to it. Like FloWM, the tracker computes the agent's own position and heading from its actions, which are exact in this environment.

\subsection{Speed snapping}
A velocity estimated from noisy positions is itself noisy, and a small velocity error grows into a large position error over a long gap. In the standard setting, every block moves at a speed of either 0 or 1 cell per step on each axis. The tracker uses this fact and snaps each axis of a track's velocity to a speed of 0 or 1. During the context phase, it snaps an axis once its velocity estimate makes one of these speeds at least 99\% probable, and it then holds that speed, reversing its sign at each bounce. When the prediction phase starts, it rounds every axis that has not yet snapped to the nearer of the two speeds. We found the two speeds from the velocities that the tracker itself estimated on 1,024 training videos of the standard setting, without true identities, and we chose the snapping thresholds on our tuning videos (App.~\ref{app:protocol}). Speed snapping is knowledge about the standard setting, in the same way that FloWM's built-in velocities are. For the varied setting we refitted the allowed speeds without labels on 1,024 varied training videos, which gives 0, 1 and 2 cells per step (App.~\ref{app:governed}).

\subsection{What the simulator gives the tracker}
For every visible block in every context frame, the tracker receives the block's true floor position with added random noise. The noise shifts the block's bearing and its distance independently, each with a standard deviation of one image pixel, so the error grows with distance as it would for a real camera. The tracker also receives the uncertainty of each noisy position. It also receives the block's true mask, and it uses the mean color inside the mask as the block's appearance. It reflects tracks off the true wall positions. At scoring time, we must also know which track describes which block. We use the true identity of each detection for this purpose, and for each block we report the track that the block's detections last matched. This readout is also an oracle step. FloWM's probe receives similar help in another form, since it is trained with the true position of each block.

\subsection{How good can the tracker be?}
\begin{table}[h]
    \centering
    \footnotesize
    \caption{Out-of-view accuracy (\%) of tracker variants on the standard test videos. The row marked ``the tracker'' is the tracker of the main text. The upper rows score all 85,602 out-of-view answers. The lower rows score only the 83,202 out-of-view answers for blocks sighted in at least two context frames. A restart predicts a block forward from the tracker's estimate at its last sighting, with no further matching.}
    \label{tab:ceiling}
    \begin{tabular}{lc}
        \toprule
        Tracker variant & Out of view \\
        \midrule
        \multicolumn{2}{l}{\emph{All blocks}} \\
        Exact positions, no speed snapping & 93.7 \\
        One-pixel noise, no speed snapping & 47.1 \\
        One-pixel noise, speed snapping (the tracker) & 90.8 \\
        Exact positions, speed snapping & 94.3 \\
        \midrule
        \multicolumn{2}{l}{\emph{Blocks sighted at least twice}\posthoc} \\
        The tracker & 93.4 \\
        Restart from the tracker's estimate at the last sighting & 94.3 \\
        Restart with the true velocity & 97.8 \\
        \bottomrule
    \end{tabular}
\end{table}

Because the tracker receives oracle inputs, its score is a ceiling for this tracker design and a reference point for a learned node memory. We first check that it is a strong reference. It keeps identities well, with an IDF1 score of 95.6\%~\citep{ristani2016idf1}. IDF1 measures how many detections a tracker assigns to the correct identity, balancing missed and wrong assignments. Table~\ref{tab:ceiling} shows how the tracker's out-of-view accuracy depends on its inputs and on speed snapping.

Speed snapping carries most of the tracker's strength under noise. With exact positions, speed snapping changes little (94.3\% against 93.7\%). Noise of a single pixel cuts this to 47.1\%, because noisy positions give noisy speeds and the errors grow over long gaps. Speed snapping restores the score to 90.8\%. These allowed speeds match the standard setting. For the varied setting we refitted them and ran the tracker once on the varied test videos (App.~\ref{app:governed}).

The lower rows of Table~\ref{tab:ceiling} show where the remaining error comes from. For each block sighted at least twice, we restart the block from the tracker's own estimate at its last sighting and predict it forward with no further matching. This restart scores 94.3\%, slightly above the tracker's 93.4\% on the same answers. The difference is error that enters after the last sighting, for example when a detection of another block is matched to the track. Giving the restart the true velocity raises it to 97.8\%, so velocity error at the last sighting costs about 3.5 points. We froze the tracker's settings on the tuning set, but only after we had seen some test-set results.

\section{Write-back}
\label{app:writeback}

During the prediction phase no block is observed, whether it is in view or not. Yet the convolutional probe scores 8 points higher on blocks in view than on blocks out of view (Table~\ref{tab:main}). One explanation would be that blocks in view were seen more recently. To test this, we compare blocks in view and out of view that have the same number of steps since their last sighting. On the test videos, FloWM keeps an in-view advantage of 7 to 10 points in every bin from 5 steps on, and every confidence interval lies above zero\posthoc. The tracker has no such advantage. Weighted to the same mix of steps since the last sighting, its in-view score minus its out-of-view score is $-$0.1 points (CI $-$1.2 to 0.9), while FloWM's is $+$7.5 points (CI 6.7 to 8.4).

The likely cause is FloWM's write-back. At each predicted step, FloWM renders a frame and writes it back into the cells in view. When a block enters the camera's view during the prediction phase, its probe accuracy rises from 84.8\% to 92.4\% on the tuning videos (read by heads trained on the other half of the tuning videos), although nothing new is observed\posthoc. On the test videos, read by the convolutional probe, it rises from 89.6\% to 94.2\% (4.6 points, CI 3.7 to 5.5), while the tracker stays at 90.5\%. We also read the memory just before each write on the tuning videos, but the probe was trained on memory read after the write and scores lower before it even on blocks that stay in view (84.1\% against 96.5\%), so this read cannot say how much of the in-view advantage the write-back causes. During prediction the written frame is computed from the memory itself, so the write adds no new information about where blocks are. In-view scores therefore mix memory with the model's own predictions, and we compare the two memories on blocks out of view.

\section{Time since the last sighting}
\label{app:recency}

Figure~\ref{fig:recency} shows how accuracy on out-of-view blocks changes with the steps since the last sighting. FloWM falls from 96\% at 2 to 4 steps to between 89\% and 90\% from 10 to 99 steps, and it ends at 82\%. The tracker starts higher, at 97\%, and falls further, to 72\%. The ranking therefore flips. Hollow markers in Fig.~\ref{fig:recency} mark bins with fewer than 1,000 answers. At 10 to 19 steps, the tracker leads by 6 points (CI 4 to 8). At 100 to 139 steps, it trails by 10 points (CI 4 to 16)\posthoc.

This flip does not by itself show that FloWM's memory fades more slowly. The late bins hold different kinds of blocks from the early bins, and two causes explain the tracker's late deficit. We found both after seeing the results\posthoc.

The first cause is blocks sighted only once. Such a block gives the tracker a position but no velocity. Its track either stays where the block was seen or, for 13 of the 15 such blocks at 100 to 139 steps, is also updated by detections of other blocks, and in either case the tracker finds almost none of them. These blocks make up 14\% of out-of-view answers at 100 to 139 steps but only 0.6\% at 10 to 19 steps. The tracker finds 0.6\% of them, while FloWM finds 30.2\%. FloWM appears to infer a block's direction from a single sighting. To check this, we took answers for blocks sighted only once in which the block's last-seen cell and the cells it would reach in each of the four directions all differ. Among FloWM's answers that land on one of the four direction cells, 89\% pick the true direction, where guessing would give 25\% (tuning videos). We have not identified the cue that FloWM uses.

The second cause is error that enters a track during the context phase after the block's last sighting, such as a match to another block's detection. For blocks sighted at least twice, we restart each block from the tracker's estimate at its last sighting, as in App.~\ref{app:tracker}. The restart beats the tracker by 6.7 points at 100 to 139 steps and by nothing at 10 to 19 steps. A long gap since the last sighting means that the context phase continued for many steps after it. During those steps, a detection of another block can be matched to the track and pull it away. Such errors do not show until the track is scored.

Splitting the gap by how often each block was sighted makes this precise. Blocks sighted once account for 4.1 of the 9.8 points (CI 0.9 to 7.9), and blocks sighted at least twice account for the other 5.7 (CI $-$0.2 to 11.7). For the latter, error that enters after the last sighting accounts for 5.8 points (CI 1.7 to 10.4), which is the 6.7-point loss against the restart weighted by these blocks' share of the bin, and the restart itself scores 0.1 points above FloWM (CI $-$4.8 to 4.6). We made this split after we had seen the results\posthoc.

\section{Varied setting}
\label{app:varied}

\subsection{FloWM on velocities outside its built-in set}
The probe scores in this subsection are overall accuracy, over answers in view and out of view. Appendix~\ref{app:governed} gives out-of-view values by motion type (Table~\ref{tab:flowmoov}).

The mix layer could, in principle, let FloWM carry more velocities than its five copies. Since the mix can move content by one cell, a block moving two cells per step could be carried by a one-cell shift in a copy plus a one-cell move in the mix. Training on the varied setting tests whether FloWM learns to do this. We compare four models. The first is the released checkpoint, which was trained on the standard setting. The second is the released checkpoint fine-tuned on the varied setting for 20k steps. The third and fourth are trained from scratch for 150k steps with FloWM's recipe, one on the varied setting and one on the standard setting. The from-scratch standard model checks how well our training run reproduces the released model.

We measure each model in two ways. The first is the MSE of predicted frames on 1,000 clips per setting, over 70 predicted frames, which needs no probe. The second is probe accuracy on the varied test videos, split by motion type into built-in, diagonal, and speed 2. For the released model, we read the varied videos with the convolutional probe trained on the standard setting, used unchanged. We also trained probe heads on the varied setting while the model generated new rollouts, rather than from saved memory states. We use such heads to read the fine-tuned and retrained models, whose memories the standard heads have never seen. Heads trained this way on the standard setting are scored on a separate set of 1,024 fresh standard videos, which include the 256 test videos.

\begin{table}[h]
    \centering
    \footnotesize
    \setlength{\tabcolsep}{2.4pt}
    \caption{FloWM on the varied setting, as overall accuracy (\%) over answers in view and out of view. The standard-video column is on 1,024 fresh standard videos, which include the 256 test videos, and every score in it comes from heads trained on the standard setting while the model generated rollouts. The varied-video columns are on the 1,024 varied test videos and use the heads named in the second column. MSE is on 1,000 clips with 70 predicted frames. \emph{Built-in} means one cell per step along one axis. Heads marked $^\ast$ were trained while the model generated rollouts, and the other heads were trained on saved memory states. Probe scores carry no interval.}
    \label{tab:s1}
    \begin{tabular}{llccccccc}
        \toprule
         & Probe & Standard & \multicolumn{4}{c}{Varied videos} & \multicolumn{2}{c}{MSE} \\
        \cmidrule(lr){4-7}\cmidrule(lr){8-9}
        Model & trained on & videos & All & Built-in & Diagonal & Speed 2 & Standard & Varied \\
        \midrule
        Released & Standard & 86.9 & 16.3 & 75.8 & 3.6 & 4.3 & 0.000649 & 0.0148 \\
        Released & Varied$^\ast$ & -- & 6.2 & 22.3 & 2.2 & 3.1 & & \\
        Fine-tuned on varied, 20k steps & Varied$^\ast$ & 84.5 & 20.1 & 46.8 & 13.4 & 15.0 & 0.00135 & 0.00747 \\
        From scratch on standard, 150k & Standard$^\ast$ & 84.8 & -- & -- & -- & -- & 0.000824 & 0.0158 \\
        From scratch on varied, 150k & Varied$^\ast$ & 30.5 & 13.1 & 20.7 & 10.5 & 11.8 & 0.00592 & 0.00790 \\
        \bottomrule
    \end{tabular}
\end{table}

For the released model, the convolutional probe trained on the standard setting, used unchanged on varied videos, finds 75.8\% of blocks that move at a built-in velocity (Table~\ref{tab:s1}). It finds only 3.6\% of diagonal blocks and 4.3\% of speed-2 blocks. On built-in-velocity blocks it scores 14 points below the 90.2\% it scores on the standard test videos (CI 13 to 16). This drop may come from the memory itself, or from memory states that the probe never saw in training. Probe heads trained on the varied setting fit their own training videos at 93\% but find only 6.2\% of blocks in fresh varied videos, so we rely on the probe trained on the standard setting for the released model. The MSE, which needs no probe, is 23 times higher on the varied setting than on the standard setting.

Fine-tuning on the varied setting for 20k steps halves the MSE on the varied setting. It also doubles the MSE on the standard setting. Its probe finds only 13\% to 15\% of diagonal and speed-2 blocks.

Training from scratch on the varied setting for 150k steps gives a varied-setting MSE only 1.33 times its own standard-setting MSE (CI 1.28 to 1.39). Its probe, however, finds only 10.5\% to 11.8\% of diagonal and speed-2 blocks, against 20.7\% of built-in-velocity blocks. This model is also weak overall. Its MSE on the varied setting is 12.2 times the released model's MSE on the standard setting. On standard clips, its MSE is 9.1 times the released model's, and its probe finds 30.5\% of blocks, against 86.9\% for the released model read by the same kind of probe. Our from-scratch run on the standard setting missed the released MSE by more than 10\% (0.000820 against 0.000603 on the release's validation clips), so our training recipe may be weaker than the original. With one seed per run, these results neither show nor rule out that the mix layer can extend the built-in velocities.

Taken together, FloWM's memory carries blocks whose motion matches its five velocities and largely loses the rest. Covering the varied setting's 24 velocities with built-in shifts would take 25 copies of the grid, one of them still, or 7.0M numbers instead of 1.4M~\citep[their Apps.~F and G.1]{lillemark2026flowm}. Training that model is out of scope for this work.

\subsection{The tracker on the varied test videos}
\label{app:governed}

We ran the tracker once on the 1,024 varied test videos, the videos on which FloWM is scored above. Before any test video was read, we froze the code and settings and registered the run and its headline number. The headline is out-of-view accuracy on diagonal and speed-2 blocks for the tracker with allowed speeds 0, 1 and 2. Each of five configurations ran once, and Table~\ref{tab:governed} reports every number as it came out. Each configuration scores 430,080 answers, of which 342,917 are out of view, and 19 blocks were never sighted in context. In-view and overall accuracy are within about half a point of the out-of-view values.

With allowed speeds 0, 1 and 2, the tracker finds 93.9\% of out-of-view diagonal and speed-2 blocks. On the same answers, FloWM's released model, read by the convolutional probe trained on the standard setting, finds 4.1\%. The paired difference is 89.8 points (CI 89.1 to 90.5). Table~\ref{tab:flowmoov} gives FloWM's out-of-view accuracy by motion type for all four rows. The other three rows saved no per-answer outputs, so their values are unpaired point estimates from accuracies logged per motion type. Pooled over diagonal and speed-2 blocks they score 2.9\%, 12.6\% and 9.1\%.

The tracker of the main text, with allowed speeds 0 and 1, keeps diagonal blocks (93.1\%) and loses speed-2 blocks (4.9\%). Per-axis speeds of 0 and 1 can express one cell per step on each axis, but not two. On speed-2 blocks this tracker does worse than the tracker with no speed snapping (59.5\%), so snapping to a wrong speed hurts more than not snapping. With exact positions, speed snapping makes no difference to the score over all blocks (97.0\% with and without it).

Three caveats apply to the headline number. The tracker receives oracle detections, and it knows the true walls. Its allowed speeds were fitted without labels on 1,024 varied training videos, but its other settings did not pass our own selection rule. Before tuning, we fixed that a setting qualifies only if, with one-pixel noise, IDF1 reaches 95\% with at most 1.1 nodes per block, and if, with exact positions, it does at least as well as no speed snapping. On 256 varied validation videos no setting passed (best IDF1 86.5\%). We therefore kept the settings chosen for the standard setting, after the varied probe-training videos had already been run with these allowed speeds.

\begin{table}[h]
    \centering
    \footnotesize
    \caption{The tracker on the 1,024 varied test videos: out-of-view accuracy (\%) by motion type, and IDF1 (\%). Each configuration ran once, after the code and settings were frozen and the run was registered.}
    \label{tab:governed}
    \setlength{\tabcolsep}{4.5pt}
    \begin{tabular}{llcccccc}
    \toprule
     & & & & & & Diagonal and & \\
    Positions & Allowed speeds & All & Built-in & Diagonal & Speed 2 & speed 2 & IDF1 \\
    \midrule
    One-pixel noise & 0, 1 and 2 & 93.4 & 91.0 & 94.1 & 93.9 & 93.9 & 93.6 \\
    One-pixel noise & 0 and 1 (the tracker) & 33.3 & 89.9 & 93.1 & 4.9 & 21.7 & 89.4 \\
    One-pixel noise & no speed snapping & 56.5 & 47.1 & 54.2 & 59.5 & 58.5 & 88.3 \\
    Exact & no speed snapping & 97.0 & 95.1 & 97.1 & 97.5 & 97.4 & 94.2 \\
    Exact & 0, 1 and 2 & 97.0 & 95.4 & 97.2 & 97.3 & 97.3 & 95.6 \\
    \bottomrule
    \end{tabular}
\end{table}

\begin{table}[h]
\centering
\footnotesize
\caption{FloWM on the 1,024 varied test videos: out-of-view accuracy (\%) by motion type, on the same answers as Table~\ref{tab:governed}. The first row comes from per-answer outputs. The other rows come from accuracies logged per motion type, and their last column assumes the first row's answer counts.}
\label{tab:flowmoov}
\begin{tabular}{llcccc}
\toprule
 & Probe & & & & Diagonal and \\
Model & trained on & Built-in & Diagonal & Speed 2 & speed 2 \\
\midrule
Released & Standard & 72.7 & 3.5 & 4.2 & 4.1 \\
Released & Varied & 18.6 & 2.1 & 3.1 & 2.9 \\
Fine-tuned on varied, 20k steps & Varied & 39.1 & 11.0 & 13.0 & 12.6 \\
From scratch on varied, 150k & Varied & 16.8 & 8.2 & 9.4 & 9.1 \\
\bottomrule
\end{tabular}
\end{table}
\section{Learned bounce}
\label{app:bounce}

The tracker's bounce is an exact reflection off the true walls, an oracle that a learned node memory will not have.
We therefore tried to learn the bounce instead, while still telling the tracker where the walls are. This experiment used the tracker with the exact reflection replaced by a learned gate, described below, and it ran on training videos of the standard setting that we generated rather than on our evaluation videos. These videos hold six to ten blocks in the same six colors, so most repeat a color, and color alone cannot tell two blocks apart. Before training, the gate leaves every velocity unchanged, so each track moves at constant velocity and unseen tracks pass through walls. On the tuning videos, this untrained variant keeps identities far worse than the tracker, with an IDF1 of 67.9\% against 95.2\%.

We added a gate that acts when a track is near a wall.
The gate multiplies the track's velocity on that axis by a learned factor $m$. A factor of $m=1$ continues, $m=0$ stops, and $m=-1$ reverses. We trained $m$ to reduce the track's position error each time a block is seen again.
We measure matching quality as the share of the tracker's matches that join a detection to a track started by another block. We changed the training objective after seeing early results, so we treat this experiment as a diagnostic\posthoc.

A tracker's motion model does two jobs. It predicts where an unseen block has gone, and it decides which track a new detection joins. Training it only to predict can therefore harm the second job. With no bounce model, the tracker matches 17.5\% of detections to the wrong block. With exact reflection off the true walls, it matches 16.3\%. With the learned bounce, it matches between 19\% and 26\%, across three learning rates and training steps 100 to 500. Learning the bounce made matching worse than having no bounce model at all.

Two errors explain this result, which we examine at a learning rate of $10^{-3}$ and step 500. First, tracks whose speed was fixed by speed snapping learn to stop at the wall rather than bounce. the median learned factor is $m=-0.006$ on the 6,375 steps where a block reaches the wall and keeps its velocity, and none of the 3,533 separate steps where it reverses gets a negative factor (median $+0.545$). When a track cannot tell whether its block will continue or reverse, stopping halfway lowers the average squared position error. However, it leaves the track behind the block, where it can capture another block's detection. Second, tracks that have drifted, and whose speed is not fixed, reverse on 89\% of the steps where the block in fact keeps going. Together, these errors raise the number of wrong-block matches from 3,928 to 5,738.

\section{Cost}
\label{app:cost}

Given oracle positions and masks, the tracker updates its memory with about $10^5$ to $10^6$ arithmetic operations per context step, by our hand count, and its state takes 14\,KB. FloWM's write of one frame takes $1.5\times10^{10}$ operations over a 5.6\,MB grid.
Of this cost, 96\% grows with the 885 memory entries it updates, which are the 177 cells in view times the five copies. This comparison flatters the tracker, because FloWM also perceives, and a deployed tracker would need a detector and a position estimator whose cost would dominate. The tracker's state also has a practical benefit. It stores each block's position and velocity as numbers, so no trained probe is needed to read them once tracks are linked to objects.

\section{Intervals and post-hoc analyses}
\label{app:posthoc}

\begin{table}[h]
    \centering
    \footnotesize
    \caption{The main-text comparisons that have an interval, in points of accuracy. Each difference is the first quantity minus the second, so a negative value means the first is lower. Intervals are 95\% bootstrap intervals.}
    \label{tab:intervals}
        \begin{tabular}{>{\raggedright\arraybackslash}p{0.62\linewidth}cc}
        \toprule
        Comparison & Difference & 95\% interval \\
        \midrule
        Standard test videos, out of view: the tracker minus FloWM read by the convolutional probe & $+$2.1 & 0.5 to 3.7 \\
        FloWM, in view minus out of view, matched on steps since the last sighting & $+$7.5 & 6.7 to 8.4 \\
        The tracker, in view minus out of view, same matching & $-$0.1 & $-$1.2 to 0.9 \\
        FloWM, a block after it enters view during prediction minus before, test videos & $+$4.6 & 3.7 to 5.5 \\
        The tracker minus FloWM, out of view, 10 to 19 steps since the last sighting & $+$6 & 4 to 8 \\
        The tracker minus FloWM, out of view, 100 to 139 steps since the last sighting & $-$10 & $-$16 to $-$4 \\
        Varied test videos, out-of-view diagonal and speed-2 blocks: the tracker with allowed speeds 0, 1 and 2 minus FloWM's released model & $+$89.8 & 89.1 to 90.5 \\
        \bottomrule
    \end{tabular}
\end{table}

Table~\ref{tab:intervals} gives the intervals that exist for comparisons in the main text. The intervals for the two causes of the tracker's late deficit are in App.~\ref{app:recency}. The other comparisons, such as the two probes, the tracker without speed snapping and the learned bounce, have none.

We made the choices marked $\dagger$ after we had seen results. The first is the split by number of sightings and what builds on it: the two causes of the tracker's late deficit, the split of that deficit by number of sightings (App.~\ref{app:recency}) and the restart rows of Table~\ref{tab:ceiling}. The second is the training objective of the learned bounce, which we changed after seeing early results (App.~\ref{app:bounce}). The third is the comparison of in-view and out-of-view accuracy at equal steps since the last sighting, which we proposed after seeing FloWM's in-view lead, together with the analysis of blocks entering view that followed it (App.~\ref{app:writeback}). The fourth is the pair of bins quoted for the flip, 10 to 19 and 100 to 139 steps, since we have no record that we chose this pair in advance. The bin edges themselves were fixed before any result by steps since the last sighting.

Four further disclosures apply. No setting of the tracker with allowed speeds 0, 1 and 2 passed our pre-set selection rule, so we kept the standard setting's settings after seeing results on varied videos outside the test set, and its run on the varied test videos was registered in advance (App.~\ref{app:governed}). We froze the tracker's settings for the standard setting on the tuning set, but only after we had seen some test-set results. A few numbers come from tuning or training videos and not from test videos: the read before the write, FloWM's choice of direction after a single sighting, the first of the two measurements of a block entering view, and the learned bounce. The FloWM values for the fine-tuned and retrained models in the lower half of Table~\ref{tab:main} are pooled from logged aggregates and carry no interval.

\end{document}